\documentclass[letterpaper, 10 pt, conference]{ieeeconf}  

\IEEEoverridecommandlockouts                              

\usepackage{graphicx}
\usepackage{amsmath}
\usepackage{amssymb}
\usepackage{xcolor,colortbl}
\usepackage{multirow}
\usepackage{comment}
\usepackage{tikz}
\usepackage{soul}

\usepackage{caption}

\newcommand{\M}[1]{\mathtt{#1}}
\newcommand{\V}[1]{\M{#1}}

\newcommand{\arr}[2]{\begin{array}{#1} #2\end{array}}
\newcommand{\mat}[2]{\left[\!\!\arr{#1}{#2}\!\!\right]}

\newcommand{\mb}[1]{\textbf{#1}}

\usepackage{url}

\DeclareMathOperator*{\argmin}{arg\,min}

\title{\LARGE \bf
Reconstruction of 3D flight trajectories from ad-hoc camera networks
}

\author{Jingtong Li$^{1\ast}$, Jesse Murray$^{1\ast}$, Dorina Ismaili$^{2}$, Konrad Schindler$^{1}$ and Cenek Albl$^{1}$%
\thanks{$^{1}$
Photogrammetry and Remote Sensing, ETH Zurich, 8093 Zurich, Switzerland {\tt\footnotesize \{firstname.lastname\}@geod.baug.ethz.ch}
}
\thanks{$^{2}$
Department of Mathematics, Technical University Munich, 85748 Garching bei München, Germany {\tt\footnotesize dorina.ismaili@tum.de}}
\thanks{$\ast$ 
Equal contribution}
}

\begin{document}

\maketitle
\thispagestyle{empty}
\pagestyle{empty}

\begin{abstract}

 We present a method to reconstruct the 3D trajectory of an airborne robotic system only from videos recorded with cameras that are unsynchronized, may feature rolling shutter distortion, and whose viewpoints are unknown. Our approach enables robust and accurate outside-in tracking of dynamically flying targets, with cheap and easy-to-deploy equipment. We show that, in spite of the weakly constrained setting,  recent developments in computer vision make it possible to reconstruct trajectories in 3D from unsynchronized, uncalibrated networks of consumer cameras, and validate the proposed method in a realistic field experiment. We make our code available along with the data, including cm-accurate groundtruth from differential GNSS navigation.

\end{abstract}

\section{Introduction}
Unmanned aerial vehicles (UAVs) are becoming ubiquitous: they are nowadays used for a wide range of tasks such as recording movies, delivering parcels, and collecting samples from inaccessible locations.
A basic capability of any mobile, airborne robotic system is to localise itself in the 3D environment. Navigation with on-board sensors (e.g., GNSS receivers, IMUs, cameras) has reached a high level of maturity, still the proliferation of UAVs calls for a method to track them \emph{outside-in}, i.e., without relying on on-board observations.
The need for external tracking arises whenever on-board sensors fail (e.g., GNSS-denied environments) or are not accurate enough (e.g., the well-known drift of visual/inertial odometry). Moreover, outside-in tracking is the only option for tracking UAVs operated by others (e.g., to enforce airspace restrictions), as we do not have access to the on-board systems.

What properties should an external tracking system for UAVs have? We suggest that, besides of course being accurate and reliable, it should also be reasonably cheap, which translates to only using standard sensors that can be mass-produced; and that it should be easy to deploy, without complicated setup and calibration procedures.

\begin{figure}[t!]
    \centering
    \includegraphics[trim={2cm 3cm 1.3cm 3cm},clip,width=\columnwidth]{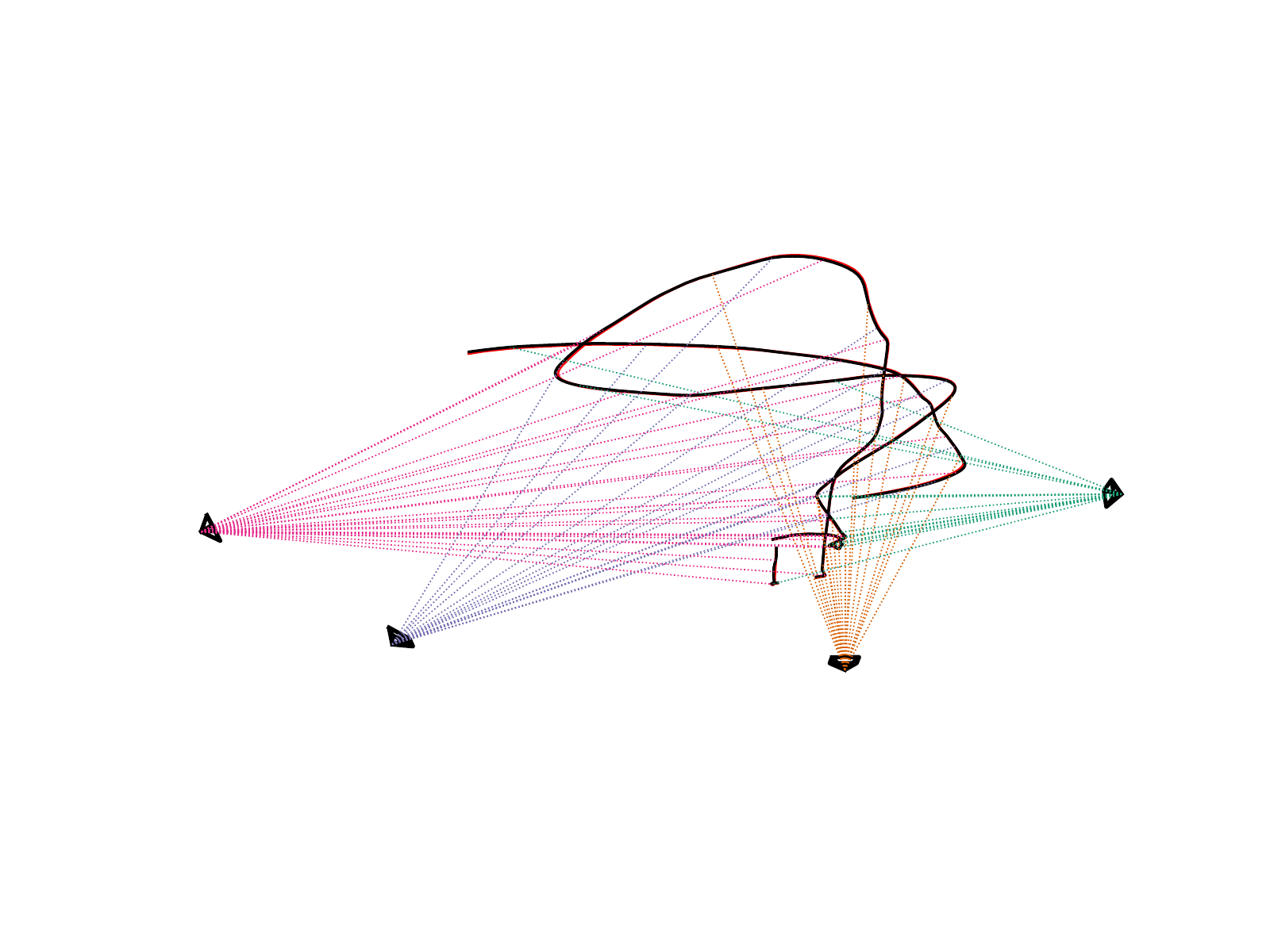}\\
    \includegraphics[width=0.49\columnwidth]{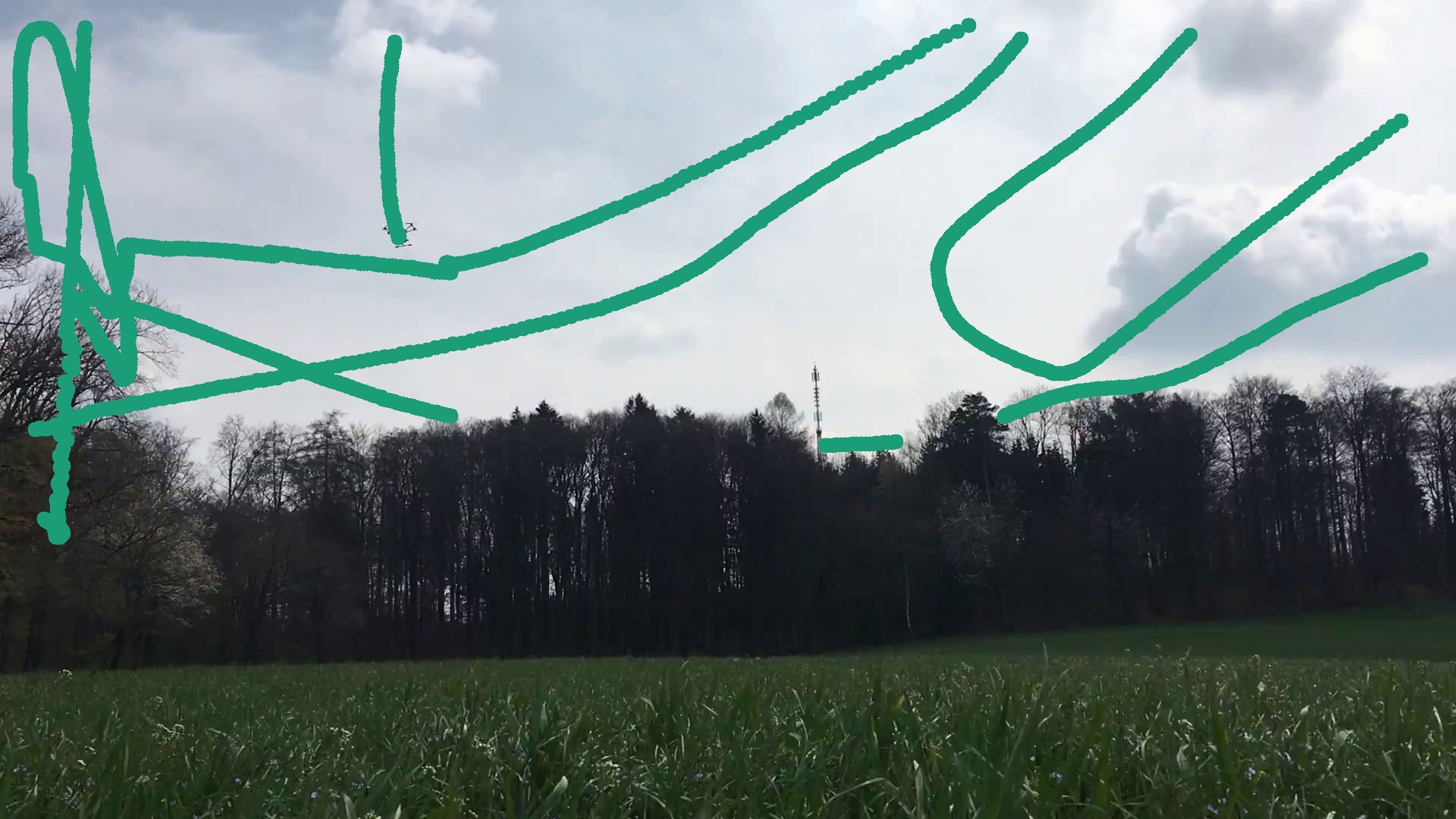}
    \includegraphics[width=0.49\columnwidth]{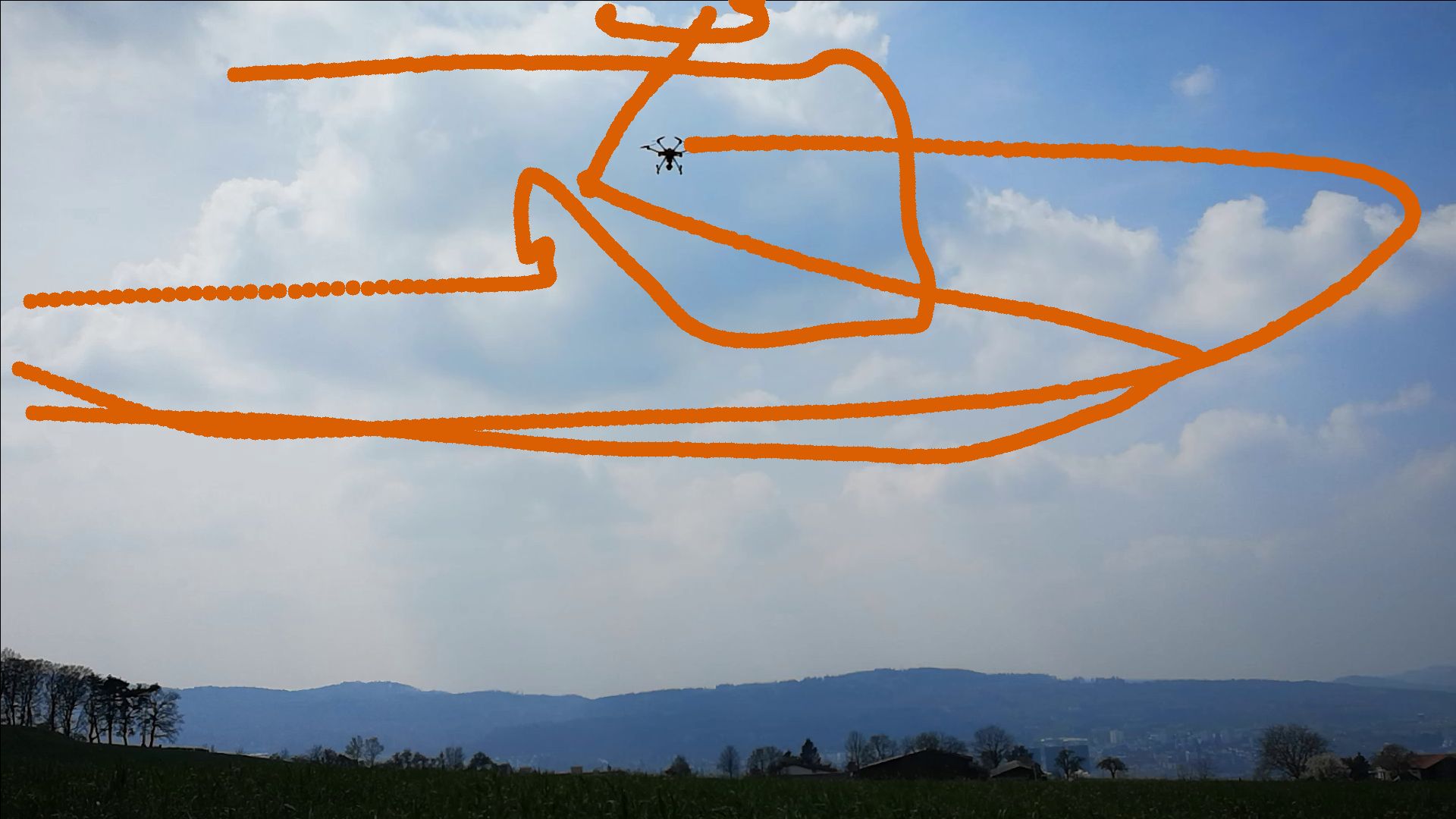}\\
    \includegraphics[width=0.49\columnwidth]{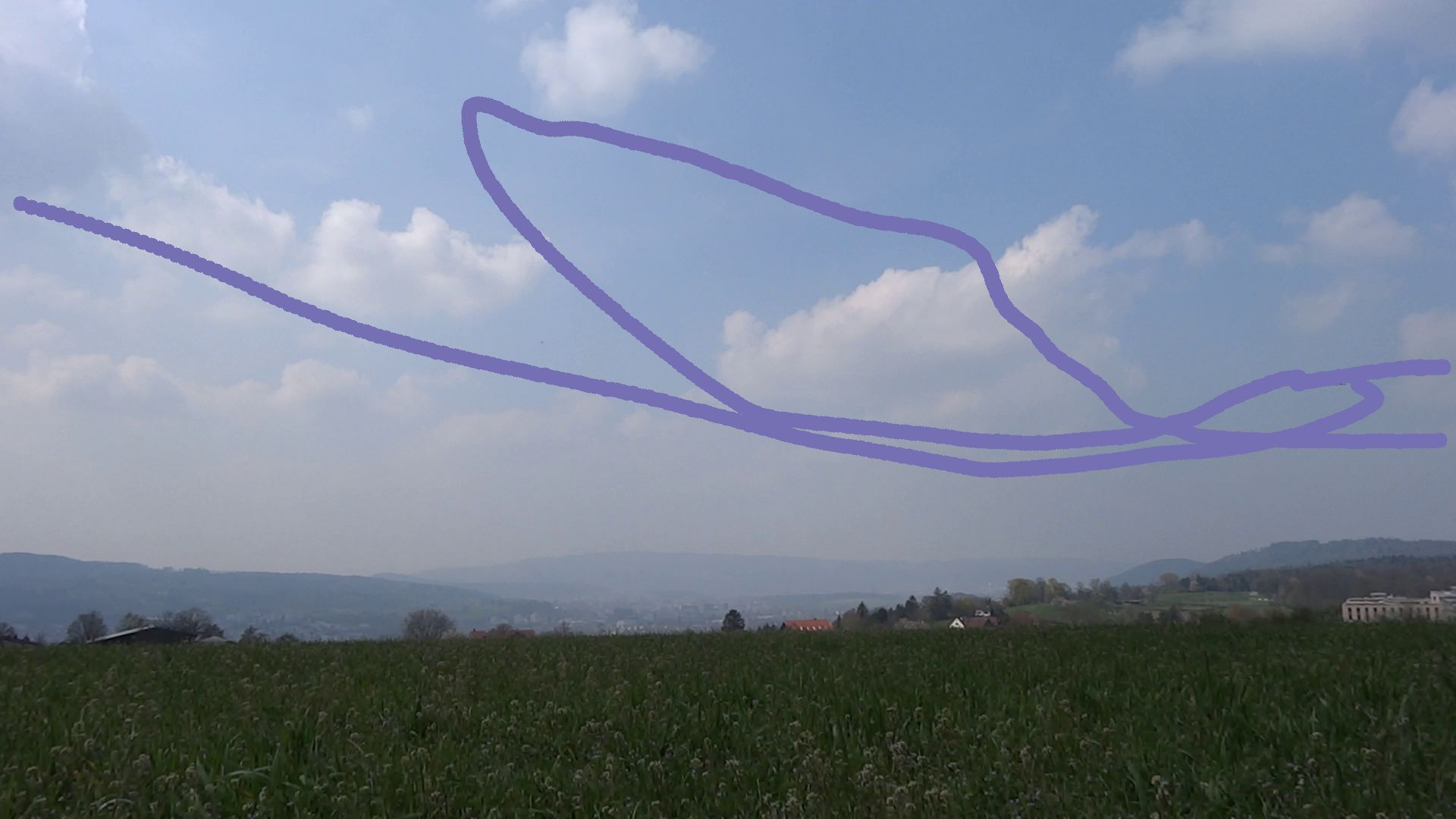}
    \includegraphics[width=0.49\columnwidth]{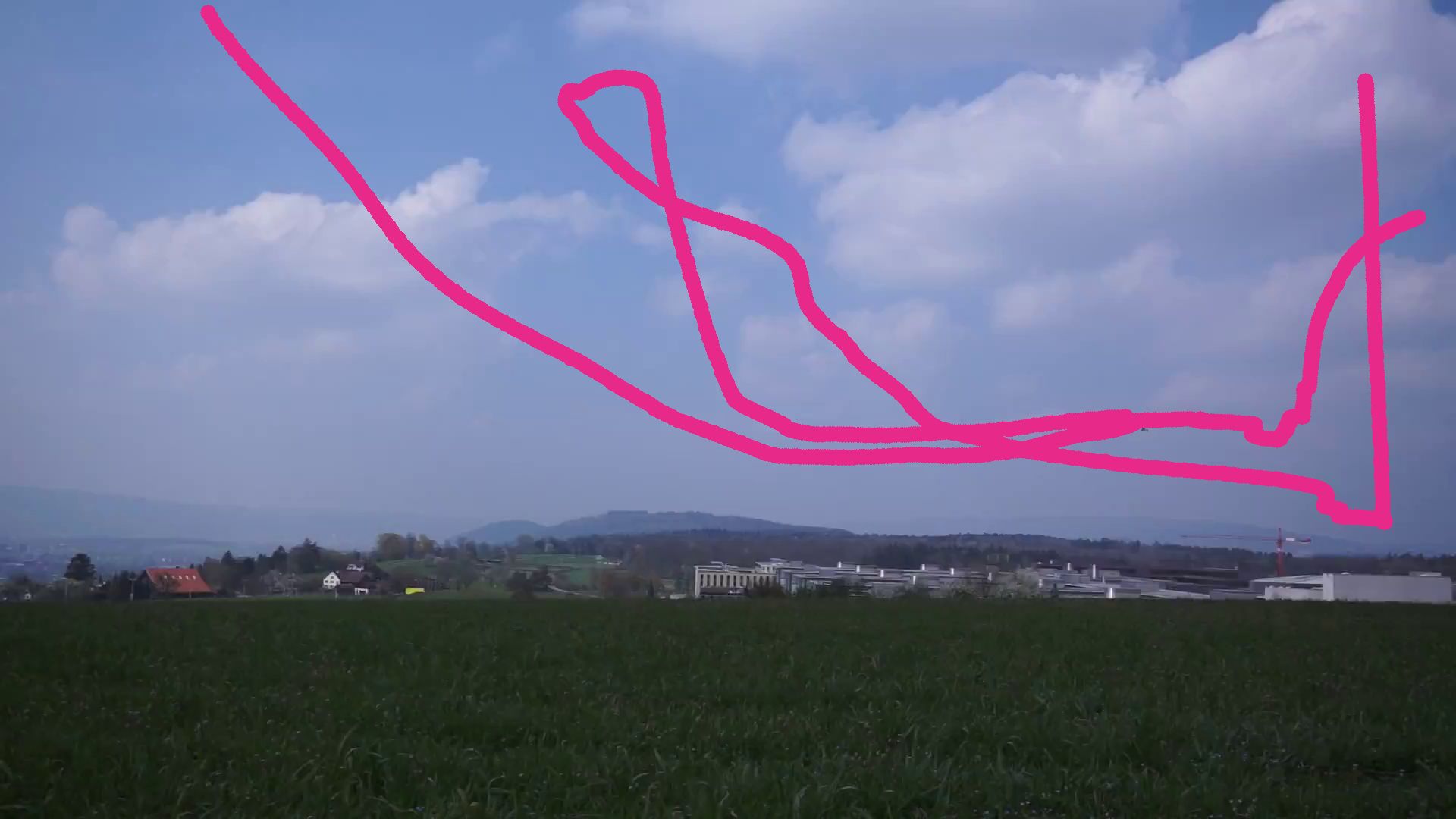}
    
    \caption{A UAV is tracked using multiple unsynchronized cameras (bottom) with unknown poses. Our method robustly retrieves the 3D trajectory (\textcolor{black}{black}), camera poses and camera synchronization. The trajectory has a mean error of 7.6 cm compared to the ground truth (\textcolor{red}{red}).}
    \label{fig:ec}
\end{figure}

Here, we propose a visual UAV tracking system consisting of a number of standard cameras (e.g., mobile phones) placed on, or near, the ground. We only assume per-camera calibration of the intrinsic parameters such as focal length and radial distortion, which can easily be done with a calibration target. All other parameters of the camera setup are unknown and can be recovered during operation, including the camera poses, the synchronisation between different cameras, and the influence of the rolling shutters. This approach allows us to use an arbitrary CMOS sensor camera network to solve problems of significantly greater complexity than standard structure-from-motion (SfM,~\cite{hartley_multiple_2003,snavely_photo_2006}) with a static scene. 

In a realistic scenario the unknown camera poses can hardly be recovered a-priori from feature correspondences, as the cameras are placed far apart and observe mostly the sky, where one can neither find matching natural image features nor place a calibration target~\cite{zhang_flexible_2000,furgale_unified_2013}. Instead, we will use the observed flying object itself as a "feature" to calibrate the camera poses.

Another challenge is that the cameras will usually not be synchronized, since hardware synchronization complicates the installation and increases the cost, especially when the distances between cameras is large. Consequently, there will be a time shift between the triggers of any two cameras, and often also a time drift due to small deviations from their nominal frame-rates. For a dynamic target, such as a UAV, that effect is significant. For instance, an offset of just 1 frame at 25 fps corresponds to 40 ms, corresponding to a displacement of 33 cm for a UAV travelling at 30 km/h. Any accuracy below that value can only be achieved with sub-frame synchronisation.

Moreover, the vast majority of cameras today are equipped with electronic rolling shutters (RS)~\cite{meingast_geometric_2005}. The sensor array is exposed line by line with a constant delay between adjacent lines, leading to time delays \emph{within} each "frame" that are proportional to the line number. Commonly the entire frame-to-frame interval is used for one RS pass, hence also  the offsets between lines in the same "frame" reach a significant magnitude for fast-moving objects.

\subsection{Previous work}
Detection and tracking of flying vehicles in images has recently been studied for a ground based, moving camera~\cite{rozantsev_detecting_2017,wu_vision-based_2017} and also for cameras mounted on another flying vehicle~\cite{nussberger_aerial_2014}. These methods pave the way for the 3D trajectory reconstruction of a flying object by detecting its image coordinates. 
A theodolite system augmented by a CCD camera, QDaedalus~\cite{guillaume_qdaedalus_2012}, has been developed to automatically and precisely measure target location. It has been since used in automatic tracking of aircraft approaching an airport~\cite{guillaume_automated_2016}. This system presents an accurate, yet very expensive method with limited applicability due to the theodolites not being suited for fast motions. 

Existing solutions for external UAV tracking use commercial motion capture systems that work with active illumination and targets placed on the UAVs~\cite{lupashin_platform_2014,how_real-time_2008,michael_grasp_2010}. 
They reach high accuracy (down to millimeters), but require precise calibration and reasonably controlled lighting; and are therefore limited to small indoor areas of at most a few tens of meters (such systems are also more expensive than many of the UAVs we may want to track).

Calibration of ad-hoc camera network has been investigated in~{\cite{devarajan_distributed_2006}} using a static scene where synchronization is not needed, contrary to the case of calibration using dynamic objects.  
In~{\cite{svoboda_convenient_2005}} it was shown that a simple moving target (laser pointer dot) is a convenient tool to accurately calibrate camera networks where the cameras are synchronized. Simultaneous calibration and synchronization from silhouettes has been proposed in~{\cite{sinha_synchronization_2004}}. That approach, while robust, suffers from high complexity, as it involves a brute-force search over possible time shifts. 

An early attempt to calibrate and synchronize cameras using the trajectory of a flying object was described in~\cite{noguchi_geometric_2007}. They employ a linear approximation of the 2D trajectory in each image to jointly estimate two-view geometry and a temporal offset. That work has been extended by~\cite{nischt_self-calibration_2009}, who use a cubic spline, and additionally estimate the difference in frame rate.
\cite{albl_two-view_2017} develops a minimal solver to estimate two-view camera geometry and time offset from eight points per view. In combination with RANSAC that method has been shown to robustly recover large offsets even in the presence of outliers.
Bundle adjustment of unsynchronized cameras observing moving objects has been investigated in~\cite{vo_spatiotemporal_2016}. The paper discusses several priors to regularise motion trajectories, and finds that, for human-induced motion, minimising the associated kinetic energy works particularly well.
\cite{rozantsev_flight_2017} present a framework for reconstructing the trajectory of a quad-rotor from ground-based cameras. They model the specific flight dynamics of their system and explicitly fit the corresponding latent motion variables like angular velocity, throttle, and moments of inertia.

\subsection{Motivation and Contribution}

Driven by the search for a convenient and practical external tracking system, this work asks the following question: can we reconstruct the 3D trajectory of a flying robot from multiple, unsynchronized videos recorded from different, unknown viewpoints?  

Our aim is to assemble a complete system
 and demonstrate it in a real outdoor experiment. Individual components of the problem have been studied~\cite{noguchi_geometric_2007,nischt_self-calibration_2009,albl_two-view_2017,vo_spatiotemporal_2016}, but they are sensitive to noise,  outliers, and temporal offsets. We are not aware that trajectory reconstruction from such an ad-hoc camera network has ever been demonstrated. 
Perhaps the closest work to ours is~\cite{rozantsev_flight_2017}, but it requires synchronized camera streams, and an existing 3D reconstruction of the scene, to which the cameras can be registered. Moreover, the method includes a dedicated motion model for quadrotors and thus cannot be used directly for other targets such as fixed-wing planes.
To our knowledge, none of the existing works has considered the temporal offsets within a frame caused by the ubiquitous rolling shutter.
  
In the following, we present a complete, fully automatic pipeline that is able to calibrate a network of independently recording cameras and to reconstruct the 3D trajectory of a flying target, using only the image locations of the target as input. While, for clarity, we base the explanation on image coordinates of the target, we point out that those were in fact determined using an off-the-shelf automatic object detection method available in OpenCV~\cite{bradski_opencv_2000} based on background subtraction and Gaussian Mixture Models~\cite{kaewtrakulpong_improved_2002}, which works well for our simple case of a UAV against a sky background.

 Our approach incrementally builds up an initial geometry estimate by chaining two-view geometries~\cite{albl_two-view_2017}, while at the same time establishing temporally consistent $2D\!\leftrightarrow\!2D$ and $2D\!\leftrightarrow\!3D$ correspondences along a spline curve. 
 The subsequent bundle adjustment~\cite{triggs_bundle_1999} is extended such that it also optimally estimates
 the temporal shifts and drifts between videos and the rolling shutter readout speeds, while constraining the 3D trajectory to a cubic spline, optionally utilizing the force/energy-based priors from~\cite{vo_spatiotemporal_2016}. Note that these priors are valid for a wide class of targets, and also simpler than a platform-specific dynamic model, as in~\cite{rozantsev_flight_2017}.
 
 The proposed method is robust against outliers, thanks to the use of a minimal solver inside a RANSAC loop for the initial geometry. Furthermore it reaches high accuracy by properly taking into account synchronization and rolling shutter. In our experiments we achieve a mean reconstruction error of $<$40 cm at $\approx$50 m flying height, using a network of four to seven different low-cost cameras (phones, compact cameras and action cams).

\section{Problem Statement}
We are interested in a broadly applicable scenario, where several different cameras are placed independently to observe the region of the sky where the UAV will operate. Note that in such a setup each camera will observe very little of the ground and baselines will be large, hence we cannot rely on background feature points for pose estimation.
The camera intrinsics are assumed to be calibrated, since this is a simple off-line procedure that can be individually performed for each camera. Recovering the intrinsics online would in principle be possible, too, but suffers from a number of geometric instabilities~\cite{wu_critical_2014,kahl_critical_2000}.
For clarity we formulate the problem for a single moving object, the extension to multiple targets is trivial.

The projection of a calibrated, perspective camera~\cite{hartley_multiple_2003} can be described as
\begin{equation}
    \V{x} = \mu(\M{P}\V{X}) = \mu(\mat{c|c}{\M{R} & \M{-C}}\V{X})
\end{equation}
where $\V{x}=\mu\mat{ccc}{x & y & z}^\top = \mat{cc}{\frac{x}{z} & \frac{y}{z}}^\top$ are the 2D image point coordinates, $\M{P}$ is the $3 \times 4$ camera matrix, $\V{X}$ are the homogeneous 3D point coordinates, $\M{R}$ is a $3 \times 3$ rotation matrix and $\V{C}$ are the 3D coordinates of the camera center.

In our network, $N$ cameras with projection matrices $\M{P}_i,i \in [0,\ldots,N-1]$ each capture $M_i$ frames at times 
\begin{equation}
t_i^j = \alpha_i j + \beta_i.    
\label{eq:time_map}
\end{equation}
Here, $\beta_i$ is an offset and $\alpha_i$ is a scale factor, which together map the frame index $j \in [0,\ldots,M_i-1]$ to a global time. Without loss of generality, we can use the first camera to anchor the global time, such that $\alpha_0 = 1$, $\beta_0 = 0$ and $t_0^j = j$.

The input of our method are the available 2D detections $\V{x}_i^j$ from all cameras. The task is to estimate the camera matrices $\M{P}_i$, synchronization parameters $[\alpha_1,\beta_1]\ldots[\alpha_{N-1},\beta_{N-1}]$, and the UAV trajectory, expressed as a list of 3D coordinates $\V{X}_i^j$, each time-stamped to a frame index $j$ in camera $i$. 

\section{Initial Geometry Computation}
As in classical SfM, respectively visual SLAM, the first step is to compute the coarse, initial camera poses and object coordinates, which are then refined through a subsequent bundle adjustment. The initial geometry is built up incrementally, by first computing the relative orientation of two views, then alternating between triangulating the 3D structure and registering new cameras with perspective-n-point (PnP).

In contrast to conventional SfM we are dealing with a dynamic object, which every camera observes at a (slightly) different time and, hence, a different 3D position. That is, we do not obtain direct $2D\!\leftrightarrow\!2D$ correspondences between images, and similarly we cannot match existing 3D points to 2D image points observed in a new camera.
Throughout our reconstruction pipeline, we use the following simple, but effective scheme.

\subsection{Correspondence between unsynchronized cameras}
\label{sec:corresp}

\begin{figure}[t]
    \centering
    \renewcommand{\tabcolsep}{1pt}
    \begin{tabular}{cc}
        \includegraphics[width=0.53\columnwidth]{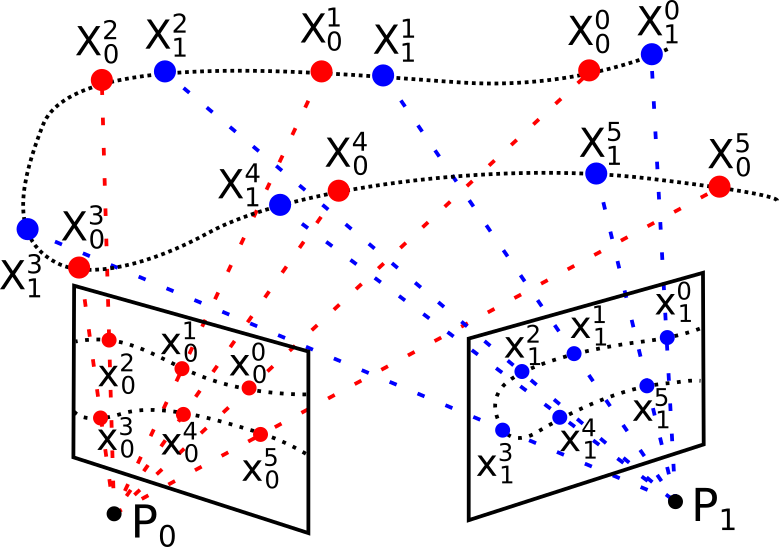} & \includegraphics[width=0.45\columnwidth]{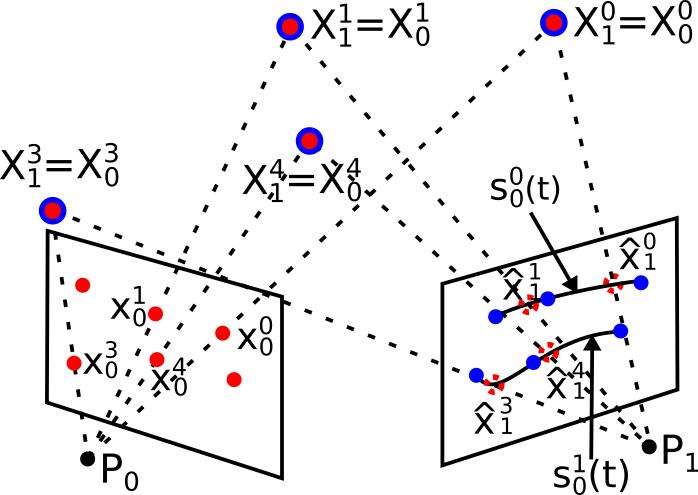} \\
        a & b\\
        \includegraphics[width=0.47\columnwidth]{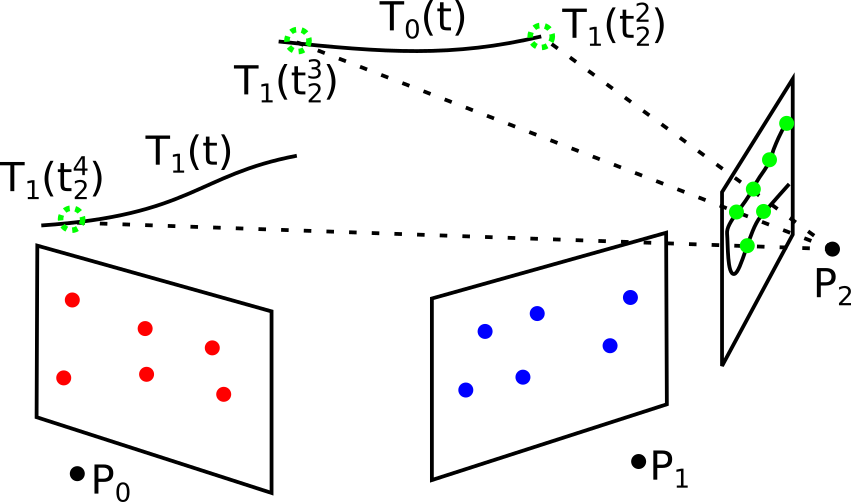} &
        \includegraphics[width=0.47\columnwidth]{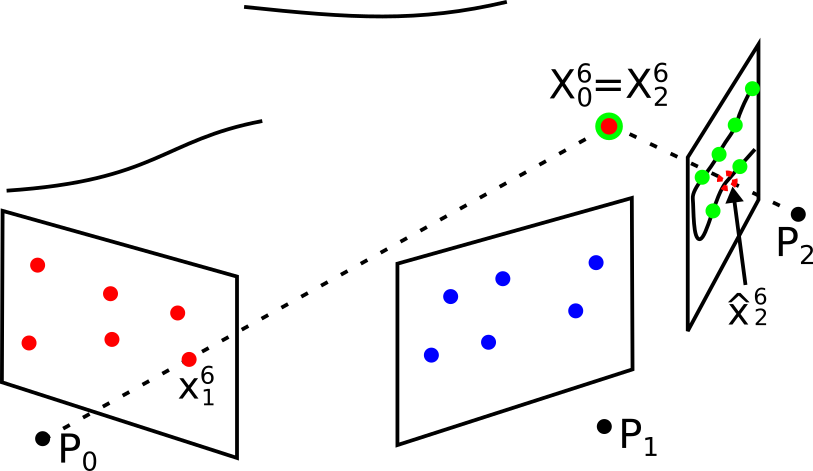} \\
        c & d \\
        \includegraphics[width=0.47\columnwidth]{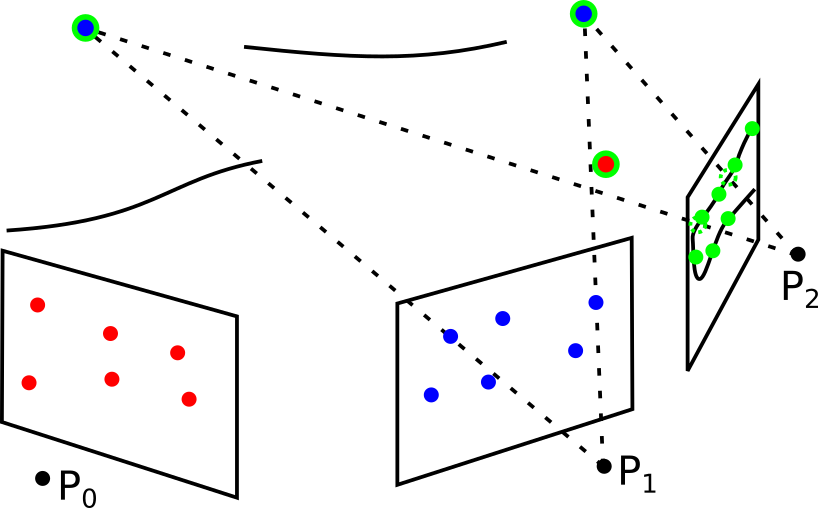} &
        \includegraphics[width=0.47\columnwidth]{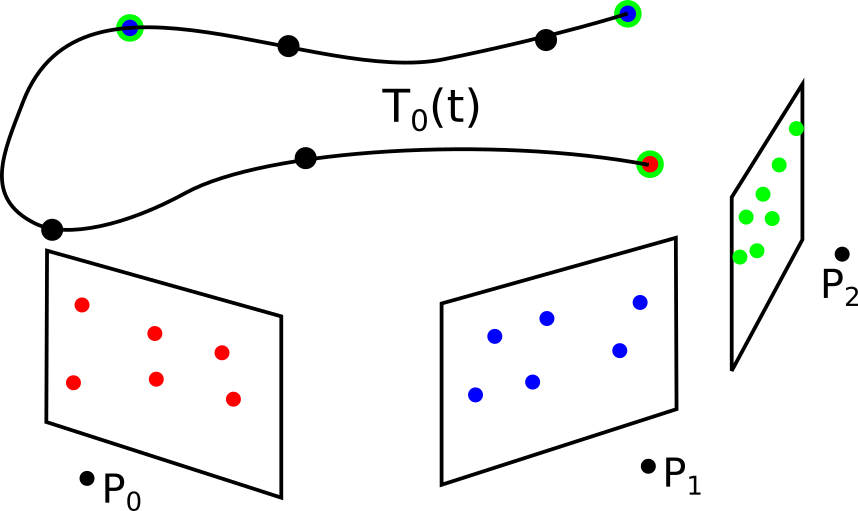} \\
        e & f \\
    \end{tabular}
    \caption{Incrementally reconstructing the trajectory. Given 2D trajectory points (a) we start with two cameras with the largest time overlap, interpolate $2D\!\leftrightarrow\!2D$ correspondences, compute relative orientation and triangulate 3D points (b). The 3D trajectory is represented by spline curves and new cameras are added by sampling the splines to obtain $3D\!\leftrightarrow\!2D$ correspondences (c). New points are triangulated by combining data from existing and newly added cameras (d,e), and the splines are updated (f).}
    \label{fig:buildingup}
\end{figure}

 To obtain $2D\!\leftrightarrow\!2D$ correspondences between two cameras $(i,k)$ we use a spline approximation of the trajectory in 2D image space. Given an image point $\V{x}_i^j$, we check whether there is a temporally overlapping set of $Q>3$ consecutive points $\{\V{x}_k^l,\V{x}_k^{(l+1)},\ldots,\V{x}_k^{(l+Q)}\}$ in the other camera. If this is the case, we fit a spline to those points,
\begin{equation}
    \V{s}^{p}_k(t)=\sum_{m}\V{b}_m(t)K^p_{km}\;,
    \label{eq:spline2D}
\end{equation}
where $\V{s}^p_k(t)$ are the coordinates on the approximated trajectory at time $t$, $\V{b}_m(t)$ are the basis functions and $K^p_{km}$ the spline coefficients. Evaluating the spline at the recording time $t_i^j$ of $\V{x}_i^j$ yields a correspondence $\V{x}_i^j\!\leftrightarrow\!\V{s}^p_k(t_i^j)$, or shortly $\V{x}_i^j\!\leftrightarrow\!\V{\hat{x}}_i^j$, see Figure~\ref{fig:buildingup}b.

$3D\!\leftrightarrow\!2D$ correspondences for registering new cameras are obtained in much the same way, by fitting the 3D trajectory with a 3D spline curve $\V{T}_r(t)$. Each $3D \leftrightarrow 2D$ correspondence is then obtained as $\V{x}_i^j \leftrightarrow \V{T}_r(t_i^j)$. See Fig.~\ref{fig:buildingup}c and Sec.~\ref{sec:trajectory} for details on 3D spline fitting.

\subsection{Two-view geometry and time shift estimation}
\label{sec:two-view}
There are two main challenges to be addressed in our scenario: outliers due to detection errors and the unknown temporal offset between the cameras. A minimal solver for fitting two-view relative pose together with a constant time offset has recently been developed~\cite{albl_two-view_2017}. Since it only needs 8 correspondences, that method is suitable for outlier rejection with RANSAC.

We obtain putative correspondences $\V{x}_i^j \!\leftrightarrow\!\V{\hat{x}}_k^j$ as described in Sec.~\ref{sec:corresp}, where as initial synchronisation parameters we set $\alpha_i$ according to the nominal frame-rate and $\beta_i$ to zero. 
We sample 8 of those correspondences randomly to solve the  generalized epipolar equation under a linear approximation $\V{v}_k^j$ of the 2D image point trajectory,
\begin{equation}
    (\V{\hat{x}}_k^j+\beta_{ik}\V{v}_k^j)^\top \M{F} \V{x}_i^j\;.
\end{equation}
This yields the relative time shift $\beta_{ik}$ and the fundamental matrix~\cite{hartley_multiple_2003} $\M{F}_{ik}$ between cameras $i$ and $k$, from which the camera matrices $\M{P}_i$ and $\M{P}_k$ can be readily extracted~\cite{hartley_multiple_2003}.

We exhaustively solve the two-view geometry for all camera pairs that have a sufficient number of correspondences, so as to obtain improved estimates of $\beta_i$ for all cameras.

\subsection{3D points triangulation}
\label{sec:triangulation}
Having recovered the pairwise relative poses and time shifts, we pick the pair with most inlier correspondences, recompute the correspondences $\V{\hat{x}}_i^j \Leftrightarrow \V{x}_k^j$ with the improved $\beta_i, \beta_k$,
and triangulate 3D points with the standard linear projective method~\cite{hartley_multiple_2003}. Note that by knowing the offsets we can globally time-stamp the  triangulated 3D points $\V{X}^j$. 
Note also, once the spline approximation of the trajectory has been computed we need not save the triangulated 3D points, as their (spline-smoothed) coordinates can be recovered from the spline parameters $\V{T}(t)$ and the global time-stamp $t$.

\subsection{Registration of remaining cameras}

Absolute camera pose is computed from $3D\!\leftrightarrow\!2D$ correspondences with the standard P3P algorithm~\cite{haralick_analysis_1991}, using RANSAC for robustness. Correspondences for a new camera $k$ are obtained by finding the temporal overlap with the already reconstructed 3D trajectory and evaluating the spline curves $\V{T}_r(t)$ at the appropriate times $t_k^j$, see Sec.~\ref{sec:corresp}. 

\subsection{Building up the trajectory}
\label{sec:trajectory}
The UAV trajectory is gradually built up as a sequence of 3D spline curves parametrized by the global time $t$, such that
\begin{equation}
    \V{T}_r (t)=\sum_{u}^{V}\V{b}_u(t)K_u
    \label{eq:spline3D}
\end{equation}
Only the spline parameters are carried over to subsequent steps of the pipeline. Every time a new camera $k$ has been registered, we try to fill in previously missing parts of the 3D trajectory. To that end the algorithm looks for image points $\V{x}_k$  outside the (temporal) range of the already reconstructed parts, constructs $2D\!\leftrightarrow\!2D$ correspondences for those points as described in Sec.~\ref{sec:corresp}, triangulates them, and extends the spline.
With a similar mechanism, existing portions of the spline are periodically updated as additional $3D\!\leftrightarrow\!2D$ correspondences become available from new images. 

\section{Rolling Shutter Effect}
\label{sec:rs}
The large majority of cameras are nowadays equipped with rolling shutters. We must account for the fact that, while the first line $l_0$ in frame $j$ is recorded at time $t_i^j$, a different line $l_k$ of the same frame is recorded at $t_i^j + r_i l_k$, with $r_i$  the RS scan speed expressed in lines/s. 

Accordingly, we incorporate the RS effect in our optimization, by modifying equation~\ref{eq:time_map} to
\begin{equation}
    t_i^j = \alpha_i j + \beta_i + r_i x_{2i}^j
    \label{eq:time_map_rs}
\end{equation}
where $x_{2i}^j$ is the $y$-coordinate (the line index) of the image coordinate, $\V{x}_i^j = \mat{cc}{x_{1i}^j & x_{2i}^j}^\top$.

\section{Spatio-temporal Bundle Adjustment}
\label{sec:BA}
As in conventional SfM, our incremental procedure is not statistically optimal and prone to error accumulation.
We thus optimize the estimated parameters by minimizing the image reprojection errors through an extended bundle adjustment~\cite{triggs_bundle_1999}.
The inputs of the optimization are the 2D observations $\V{x}_i^j$ and initial estimates for all unknown parameters, as obtained from the initial reconstruction described above.
That is, the camera poses $\M{P}_i$, 3D trajectory parameters $T_r(t)$, as well as time offsets $\beta_1,\ldots,\beta_n$, time scales $\alpha_1,\ldots,\alpha_n$ and  RS scan speeds $r_i$ for each camera. For the latter, initialising them as $r_i=0$ is sufficient.

The splines used so far are a convenient, but somewhat arbitrary choice of trajectory model. For the refinement we therefore also experiment with other ways of regularizing the trajectory. Namely, minimizing respectively the kinetic energy and force associated with the UAV motion.

\subsection{Spline trajectory}
This variant simply retains the cubic splines $T_r(t)$ already fitted before, and optimizes their parameters $K_r$:
\begin{equation}
\begin{split}
    &\argmin_{\M{P}_i,\alpha_i,\beta_i, r_i, K_r} \sum_{i=0}^{N}{\sum_{j=0}^{M_i}{\Vert\V{x_i^j}-
    \mu(\M{P}_i T_r(t_i^j))\Vert^2}}
    \\
    &\text{with}\quad t_i^j = \alpha_i j + \beta_i + r_i x_{2i}^j\;.
    \end{split}
    \label{eq:opt_spline}
\end{equation}

\subsection{Least kinetic energy trajectory}
\label{sec:least_ke}
This model was found in~\cite{vo_spatiotemporal_2016} to work well for dynamic 3D objects. The kinetic energy of an object with mass $m$ is $E_k=\frac{1}{2}m v^2$. Considering the object a point mass with $m=1$, and approximating its instantaneous velocity at time $t_i^j$ as $||(\V{X}_i^{j+1}-\V{X}_i^j)/(t_i^{j+1}-t_i^j)||$\, leads to
\begin{equation}
\begin{split}
    &\argmin_{\M{P}_i,\alpha_i,\beta_i, r_i, \V{X}_i^j}
    e_r+e_m\qquad\text{with}\\
    &e_r=\sum_{i=0}^{N}{\sum_{j=0}^{M_i}{\Vert\V{x_i^j}-\mu(\M{P}_i\V{X}_i^j)\Vert^2}}\\
    &e_m = \sum_{i=0}^{N}{\sum_{j=0}^{M_i}{\left\Vert\frac{\V{X}_i^{j+1}-\V{X}_i^j}{t_i^{j+1}-t_i^j}\right\Vert^2}}\;.
     \end{split}
    \label{eq:opt_leastkinetic}
\end{equation}
The times $t_i^j$ are again expanded as in equation~(\ref{eq:time_map_rs}). In our experiments, however, we found that constraining the motion purely based on the instantaneous velocity does not work at all for the problem of UAV trajectory reconstruction. The reason for this is that contrary to ~\cite{vo_spatiotemporal_2016}, who used abundant static features visible in their scenes to constrain their camera poses, the UAV scenes contain mostly uniform sky from which few static features can be identified. 

Therefore, we decided to keep the spline approximation and add the motion regularizers on top of it, such that 
\begin{equation}
    e_r=\sum_{i=0}^{N}{\sum_{j=0}^{M_i}{\Vert\V{x_i^j}-
    \mu(\M{P}_i T_r(t_i^j))\Vert^2}}
    \label{eq:opt_leastkinetic_withspline}
\end{equation}
as in equation~\ref{eq:opt_spline}.
\subsection{Least force trajectory}
This variant was found in~\cite{vo_spatiotemporal_2016} to perform similar to least kinetic energy, for human motion. For UAVs, least force is arguably a physically more meaningful assumption: the least kinetic energy prior penalizes high velocities, but for a flying robot maintaining even a rather high velocity may be less effort than applying a high force to accelerate or decelerate.

The force needed to accelerate an object with mass $m$ by $a$ is $F=ma$. Again setting $m=1$, without loss of generality, the instantaneous acceleration is approximately $a = ||(\V{X}_i^{j+1}-\V{X}_i^j)/(t_i^{j+1}-t_i^j) - (\V{X}_i^{j}-\V{X}_i^{j-1})/(t_i^{j}-t_i^{j-1})||$. The corresponding objective is the same as in section~\ref{sec:least_ke}, except that the prior is replaced by
\begin{equation}
    e_m = \sum_{i=0}^{N}{\sum_{j=0}^{M_i}{\left\Vert\frac{\V{X}_i^{j+1}-\V{X}_i^j}{t_i^{j+1}-t_i^j} - \frac{\V{X}_i^{j}-\V{X}_i^{j-1}}{t_i^{j}-t_i^{j-1}}\right\Vert}}\;,
    \label{eq:force_term}
\end{equation}
with the times $t_i^j$ again expanded as above.
The bundle adjustment can be performed periodically in the pipeline and as early as we obtain the first geometry estimate from two cameras. We perform the optimization every time a new camera has been added.

\section{Implementation Strategies}

The proposed framework allows for different modes of operation, according to application requirements.
A minimal requirement is that the observed trajectory is long enough, and deviates sufficiently from a straight line, the latter being a degenerate configuration~\cite{hartley_critical_2006}.
For offline processing, where the trajectory reconstruction is done after recording, one can simply use all collected data (subject to memory limits).
On the contrary, in an online application one will first use a limited number of detections/frames per camera to calibrate the system. Once the camera parameters have stabilized one can freeze them and switch to tracking mode, where one only triangulates new 3D points to extend the trajectory, as described in Sec.~\ref{sec:triangulation}.
As in many SfM and visual SLAM systems~\cite{snavely_photo_2006, klein_parallel_2007}, bundle adjustment (Sec.~\ref{sec:BA}) can be re-run periodically in an asynchronous thread when a sufficient amount of new evidence has accumulated, in both calibration and tracking mode.

\section{Experiments}

We use a Yuneec hexacopter designed for cinematographic recording and photogrammetric mapping. To acquire ground truth, the UAV was equipped with a real-time kinematic (RTK) GNSS system from Fixposition~\cite{noauthor_fixposition._2017}, with an estimated localization accuracy of $\pm$1 cm~\cite{su_zhenzhong_single-frequency_2018}. The UAV flies within a region of $\approx$100$\times$100 m, at heights up to $\approx$50 m.

Around that region we set up 4-7 cameras, in such a way that the UAV is almost always visible in $\geq$2 views.
The cameras are of various types -- smartphones, compact cameras and a GoPro action cam -- with varying resolutions, frame rates and fields of view, see Table~\ref{tab:cameras}. All of them have a rolling shutter.
For dataset 3, ground truth camera synchronization was obtained with a radio-synchronized network of LEDs. One LED was placed in the field of view of each camera, such that synchronous flashes are periodically visible in all videos (40 flashes in total). To evaluate also the estimated camera poses, we measured the camera locations for dataset 3 directly with a survey-grade GNSS receiver (Trimble R8). The estimated positioning accuracy is 9 mm, but since we could not make the antenna centre coincide with the projection centres of the different consumer devices, we conservatively estimate that the ground truth coordinates of the camera centres have an accuracy better than 5 cm.

Altogether, we created 4 datasets with various properties, see table~\ref{tab:datasets}. Datasets 1 and 2 represent standard cases, where the UAV flies with moderate speed and changes direction smoothly. Dataset 3 poses greater challenges in terms of speed (and thus also trajectory length) and acceleration. We strove to cover a variety of flight modes and movement types, from straight stretches to sharp changes of direction and fast ascents/descents. Dataset 4 is even more challenging, with the most aggressive flight path and also a larger number of misdetections, caused by fast-moving clouds that interfered with our detector. Dataset 5 was provided by~\cite{rozantsev_flight_2017}. Unfortunately its ground truth accuracy is not known. Off-the-shelf systems without differential GNSS typically reach at best 0.5 m, so the results cannot be compared directly to those of datasets 1-4. We refrain from any preprocessing or filtering of the videos to test that all parts of our pipeline are robust to outliers.  

\begin{table}[tb]
    \centering
    \begin{tabular}{c|c|c|c}
        Camera \# & \textbf{Model} & \textbf{Resolution} & \textbf{Frame rate} \\
        \hline
        1 & Huawei Mate 7 & 1920x1080 & 30.02 \\
        2 & Huawei Mate 10 & 3840x2160 & 29.72 \\
        3 & Sony NEX-5N & 1440x1080 & 25 \\
        4 & Sony Alpha 5100 & 1920x1080 & 29.97 \\
        5 & Sony DSC-HX20V & 1920x1080 & 50 \\
        6 & GoPro 3 & 1920x1080 & 59.94 \\
        7 & Huawei P20 Pro & 3840x2160 & 29.83
    \end{tabular}
    \caption{Cameras used to record our datasets.}
    \label{tab:cameras}
\end{table}

\begin{table}[tb]
    \centering
    \begin{tabular}{c|c|c|c|c|c}
        Dataset \# & \textbf{\# cameras} & \textbf{duration [s]} & \textbf{source} & \multicolumn{2}{c}{\textbf{Velocity [km/h]}} \\
        & & & & Mean & Max \\
        \hline
        1 & 4 & 120 & ours & 7 & 21\\
        2 & 4 & 120 & ours & 14 & 26\\
        3 & 6 & 600 & ours & 20 & 35\\
        4 & 7 & 600 & ours & 24 & 37\\
        5 & 6 & 500 & \cite{rozantsev_flight_2017} & 12 & 29
    \end{tabular}
    \caption{Datasets and their parameters.}
    \label{tab:datasets}
\end{table}


\begin{table}[tb]
    \centering
    \renewcommand{\tabcolsep}{2pt}
    \begin{tabular}{l|l|c|c|c|c|c}
        & $\beta_{init} $  & 5 & 10 & 15 & 20 & 50 \\
        \hline
         \multirow{3}{*}{Full pipeline} & $\beta_{F}$ error & 0.6 & 0.7 & 0.7 & 0.8 & 0.7  \\\cline{2-7} 
         & $\beta_{opt}$ error & 0.5 & 0.6 & 0.6 & 0.7 & 0.6  \\\cline{2-7}
         & Mean err. [cm] & 17.3 & 17.1 & 16.8 & 16.4 & 16.3 \\
         \hline
         \multirow{2}{*}{w/o $\beta_{F}$} & $\beta_{opt}$ error & 1 &  & 7.7 &  &   \\\cline{2-7} 
         & Mean err. [cm] & 19.7 &  & 83.8 &  &  \\
         \hline
         w/o $\beta_{F}$, w/o $\beta_{opt}$ & Mean err. [cm] & 137.4 &  &  &  & 
    \end{tabular}
    \caption{Importance of pairwise synchronisation. The table shows the mean error of estimated time shifts (unit: frames) for increasing initial de-sync $\beta_{init}$. Missing numbers indicate failures that did not return a reasonable trajectory. See text for details.}
    \label{tab:results_sync}
\end{table}

\begin{table}[tb]
    \centering
    \begin{tabular}{c|l|cc|cc|cc}
          \multicolumn{2}{c|}{} & \multicolumn{2}{c|}{Spline} & \multicolumn{2}{c|}{Least kinetic} & \multicolumn{2}{c}{Least force} \\
        \multicolumn{2}{c|}{} & w/o RS & RS & w/o RS & RS & w/o RS & RS \\
       \hline
        \multirow{4}{*}{\rotatebox{90}{Dataset 1}} & Mean & 7.6 & 7.5 & 7.5 & 7.4 & 7.5 & \mb{7.3}  \\
        & RMSE & 8.7 & 8.6 & 8.6 & \mb{8.5} & 8.7 & \mb{8.5}  \\
        & Median & 6.3 & 6.3 & 6.2 & 6.2 & 6.3 & \mb{6.1}  \\
        & Outl. \% & 0.0 & 0.0 & 0.0 & 0.0 & 0.0 & 0.0  \\
        \hline
        \multirow{4}{*}{\rotatebox{90}{Dataset 2}} & Mean & 14.4 & 14.3 & 14.2 & \mb{14.1} & 14.3 & 14.2  \\
        & RMSE & 16.9 & 16.8 & 16.7 & \mb{16.5} & 16.9 & 16.7  \\
        & Median & 12.1 & 12.2 & \mb{12.0} & 12.1 & 12.2 & 12.3  \\
        & Outl. \% & 0.0 & 0.0 & 0.0 & 0.0 & 0.0 & 0.0  \\
        \hline
        \multirow{4}{*}{\rotatebox{90}{Dataset 3}} & Mean & 16.8 & 16.2 & 16.6 & \mb{16.1} & 16.6 & 16.4  \\
        & RMSE & 24.5 & 22.2 & 23.7 & \mb{22.0} & 22.7 & \mb{22.0}  \\
        & Median & 11.4 & 11.5 & \mb{11.3} & 11.6 & 11.8 & 12.1  \\
        & Outl. \% & 2.2 & 1.9 & 2.2 & 1.8 & 1.8 & \mb{1.7}  \\
        \hline
        \multirow{4}{*}{\rotatebox{90}{Dataset 4}} & Mean & 38.9 & 38.2 & 38.6 & 38.0 & 37.6 & \mb{35.9}  \\
        & RMSE & 54.1 & 54.0 & 57.0 & 54.5 & 54.4 & \mb{49.7}  \\
        & Median & 31.7 & 30.2 & 29.8 & 30.1 & \mb{29.6} & 29.7  \\
        & Outl. \% & \mb{1.5} & \mb{1.5} & 1.8 & 1.6 & \mb{1.5} & \mb{1.5}  \\
    \end{tabular}
    \caption{3D trajectory errors on our datasets with accurate ground truth, in centimeters.}
    \label{tab:results_traj_ours}
\end{table}

\begin{table}[tb]
    \centering
    \begin{tabular}{c|l|cc|cc|cc}
          \multicolumn{2}{c|}{} & \multicolumn{2}{c|}{Spline} & \multicolumn{2}{c|}{Least kinetic} & \multicolumn{2}{c}{Least force} \\
        \multicolumn{2}{c|}{} & w/o RS & RS & w/o RS & RS & w/o RS & RS \\
        \hline
        \multirow{4}{*}{\rotatebox{90}{Dataset 5}} & Mean & 77.2 & 75.2 & 76.5 & 74.5 & \mb{70.5} & 73.0  \\
        & RMSE & 94.4 & 88.2 & 94.9 & 87.8 & \mb{81.7} & 85.8  \\
        & Median & 66.8 & 64.5 & \mb{60.5} & 61.1 & \mb{60.5} & 61.9  \\
        & Outl. \% & 0.8 & 0.2 & 1.4 & \mb{0.0} & \mb{0.0} & \mb{0.0}  \\
    \end{tabular}
    \caption{3D trajectory errors on the dataset of~\cite{rozantsev_flight_2017}, in centimeters.}
    \label{tab:results_traj_ms}
\end{table}

\subsection{Reconstruction accuracy}
Our pipeline achieves remarkable accuracy for both the reconstructed trajectory and the camera poses. Quantitative evaluations of the 3D error are given in Table~\ref{tab:results_traj_ours} for datasets 1-4, and in Table~\ref{tab:results_traj_ms} in dataset 5.  
To compute the error we follow~\cite{rozantsev_flight_2017}: we first fit a similarity transform to align the reconstructed trajectory to the ground truth and compensate possible systematic offsets, then compute distances between the aligned trajectories.
For dataset 3 we also measured the ground truth camera locations, which we again align to the estimated ones.
On top of the mean and median absolute errors and the RMSE, we also show the outlier ratio, computed as the portion of the trajectory outside the confidence interval of 3$\times$RMSE (99.7\% of the probability mass for 0-mean Gaussian inliers).

\begin{figure*}[tb]
    \centering
    \includegraphics[width=0.97\columnwidth]{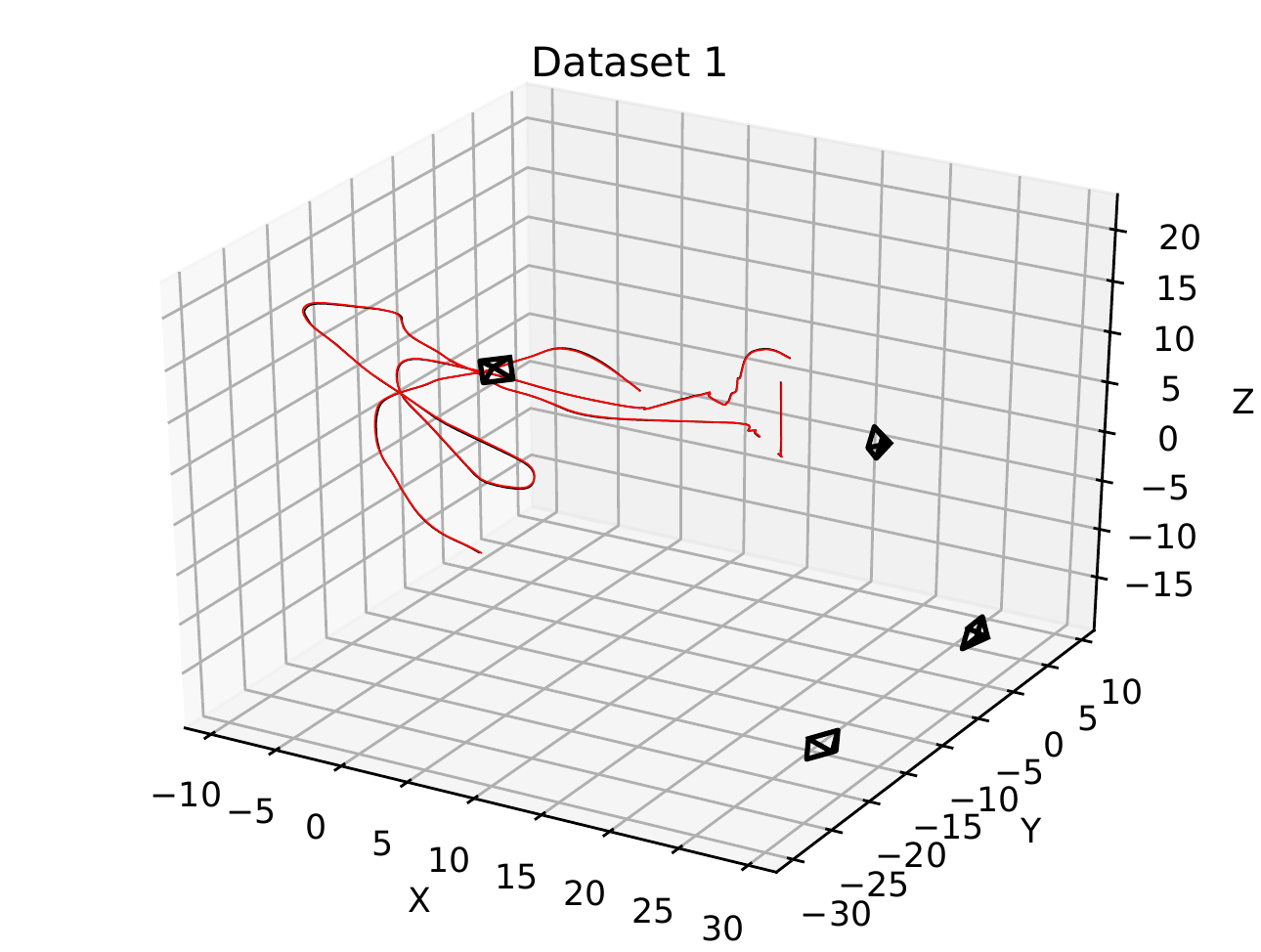}
    \includegraphics[width=0.97\columnwidth]{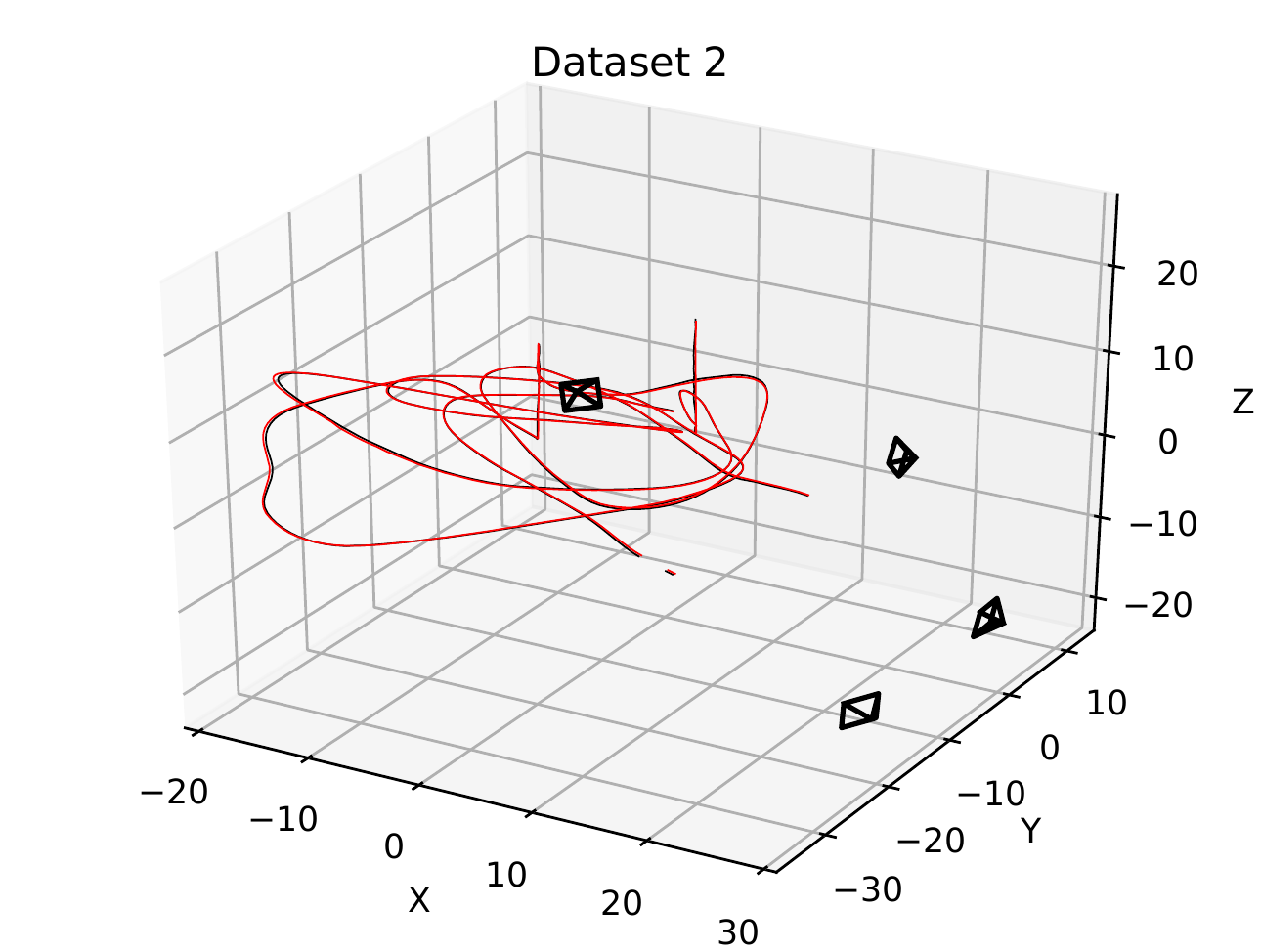}\\
    \includegraphics[width=0.97\columnwidth]{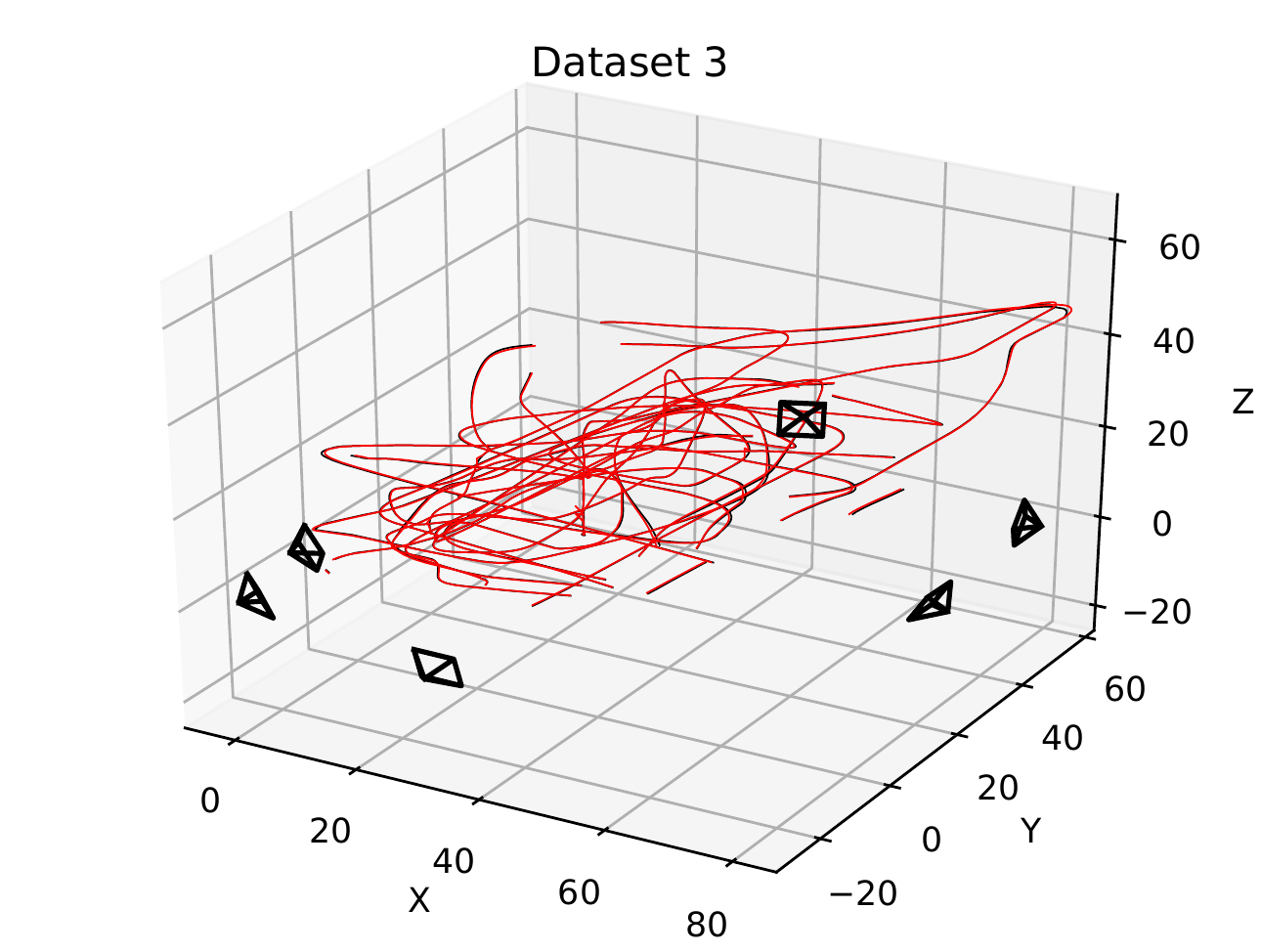}
    \includegraphics[width=0.97\columnwidth]{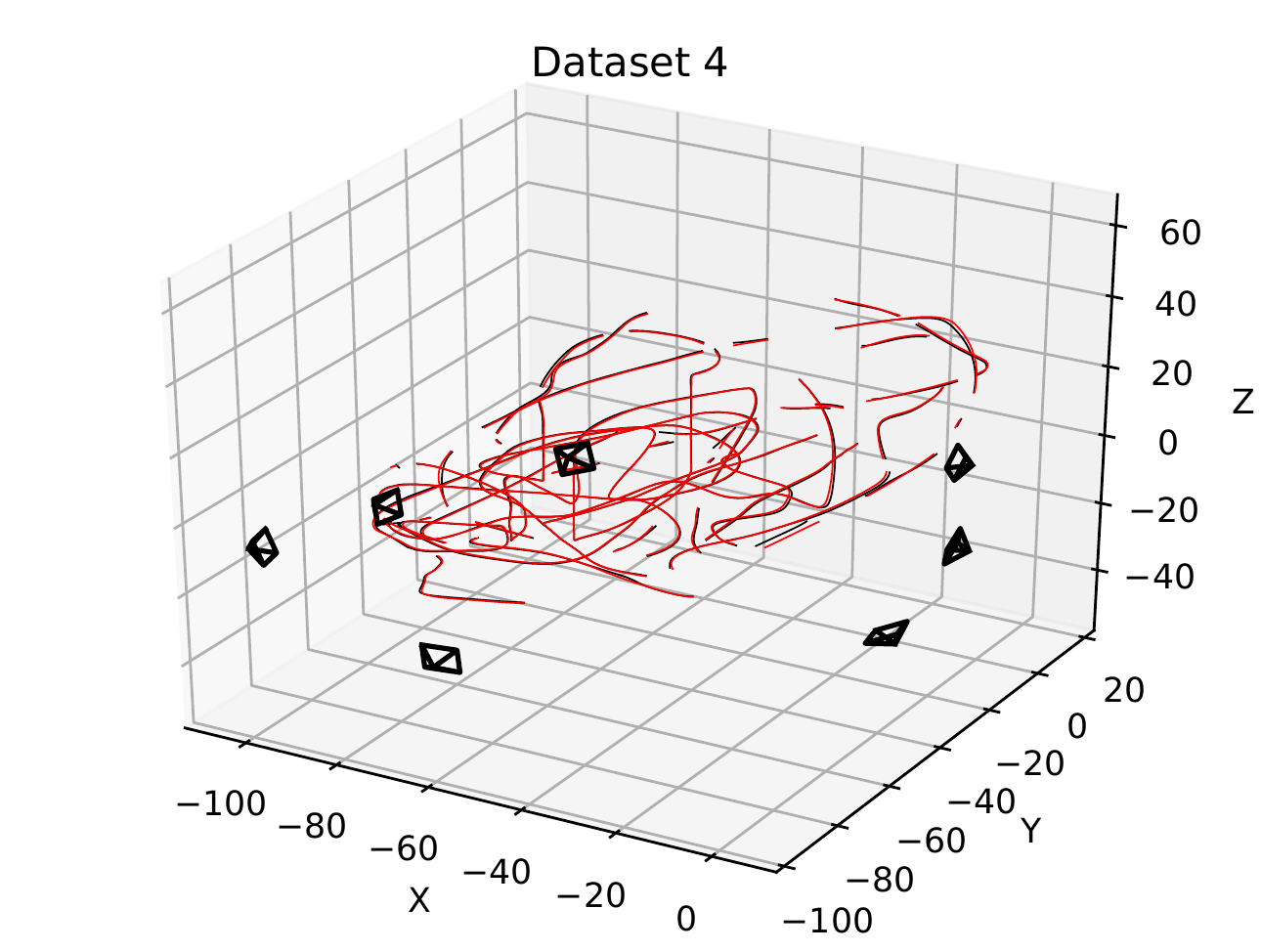}
    \caption{Visualization of the reconstructed trajectories (black) from our datasets with accurate ground truth (red). The missing parts of the trajectory are where the drone was detected in $\leq$ 1 camera.}
    \label{fig:visualization_ours}
\end{figure*}

On our data we achieved mean errors well below 20 cm for the first three datasets, and even for the hard dataset 4 below 40 cm (about the diameter of the UAV).
For dataset 3 the mean error of the reconstructed camera positions was 17cm and the biggest error was 68cm, a very good localisation accuracy considering the baselines in the order of 100 m.
Visual example of the camera poses and trajectories are shown in Fig.~~\ref{fig:visualization_ours}.

On dataset 5 we obtained an RMSE slightly above 80 cm, about half the error reported in~\cite{rozantsev_flight_2017}; a significant improvement, especially taking into account that in~\cite{rozantsev_flight_2017} the camera poses are pre-calibrated and synchronised, and the method is restricted to a specific type of aircraft dynamics.
We do point out that the comparison is not exact, because we only reconstructed approximately 90\% of the trajectory (our current detector assumes a sky background and missed the start and end of the trajectory where the UAV is near the ground).

\begin{figure}
    \centering
    \begin{tikzpicture}
        \node[anchor=south west,inner sep=0] at (0,0) {\includegraphics[width=\columnwidth]{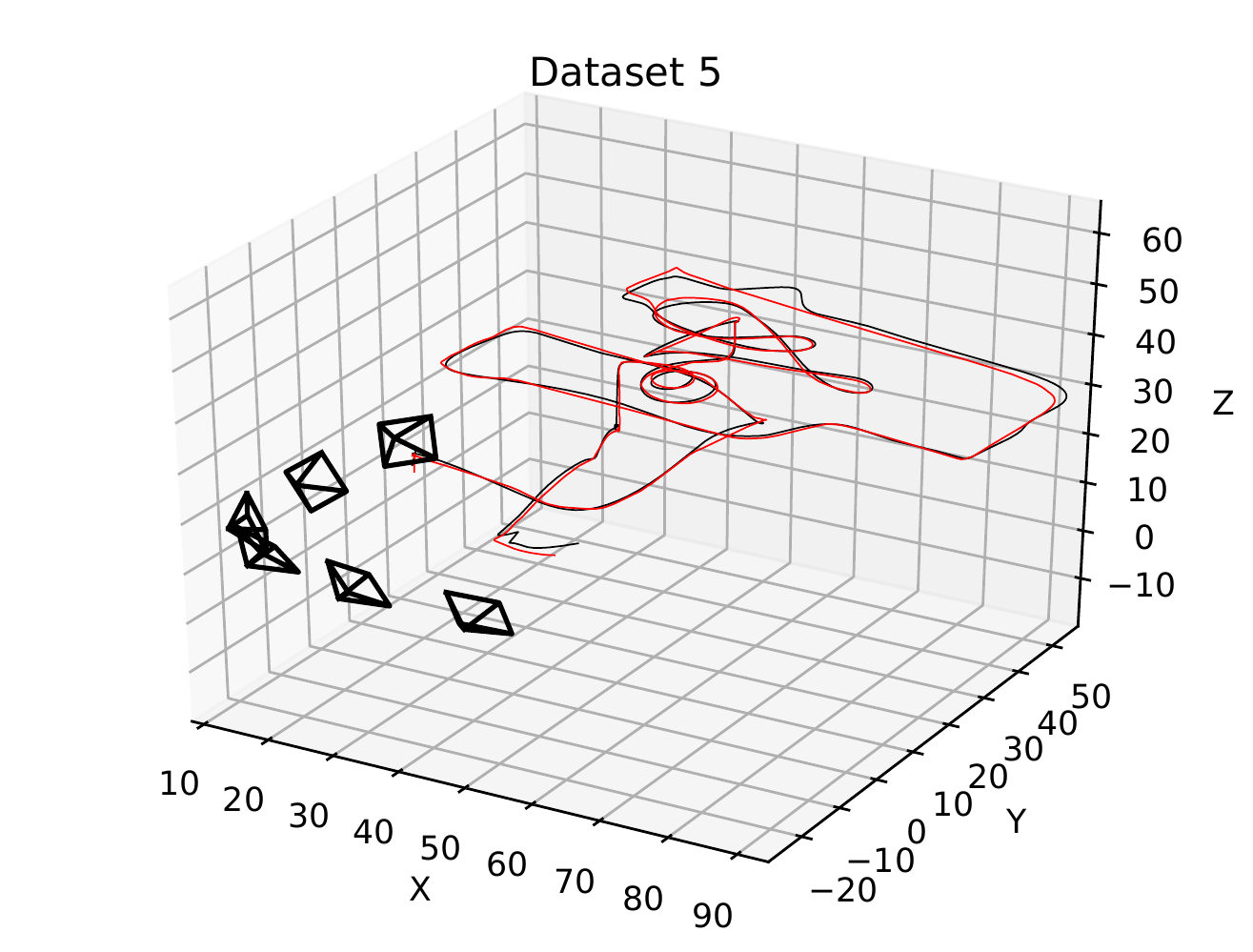}};
        \draw[blue,thick] (5.4,4.5) circle (0.3cm);
        \draw[blue,thick] (3.7,2.8) circle (0.3cm);
    \end{tikzpicture}
    \begin{tikzpicture}
        \node[anchor=south west,inner sep=0] at (0,0) {\includegraphics[width=\columnwidth]{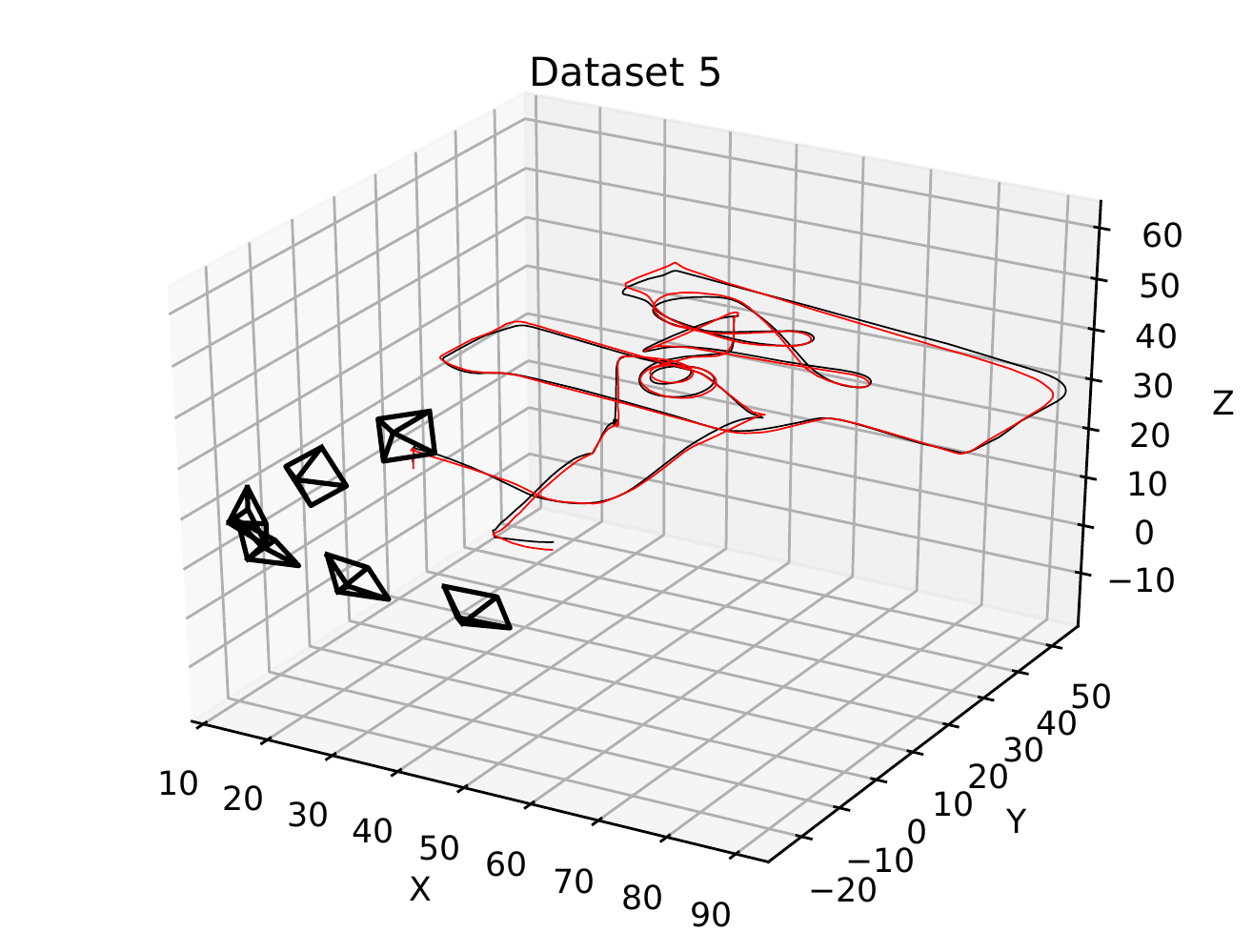}};
        \draw[blue,thick] (5.4,4.5) circle (0.3cm);
        \draw[blue,thick] (3.7,2.8) circle (0.3cm);
    \end{tikzpicture}
    \caption{Reconstructed trajectory from the data of~\cite{rozantsev_flight_2017}. Note that the spline parameterization (top) itself exhibited some artefacts (in blue circles), which the least force regularizer (bottom) was able to remove.}
    \label{fig:dataset5_diff}
\end{figure}

\subsection{Contribution of motion regularizers and RS}
Overall, the different versions of trajectory regularization did not differ significantly in terms of trajectory and camera pose errors, see Table~\ref{tab:results_traj_ours}. For the easy datasets 1 and 2, the results were almost identical. For the more challenging datasets, a small improvement becomes apparent. On dataset 3, the RMSE drops by 1 cm with the kinetic energy regularizer and by 2 cm with the least force regularizer. The number of outliers was also reduced from 2.2\% to 1.8\%. On dataset 4, the kinetic energy regularizer performed slightly worse in terms of RMSE than the spline, which could be attributed to the fact that the UAV speed was the highest in this dataset and imposing constraints on the velocity is counterproductive. On the other hand, least force prior together with the RS term improved the mean error by 3 cm and RMSE by almost 5 cm compared to other approaches.  

Incorporating the RS model in the bundle adjustment provides negligible error reduction in the easy cases, but does on average bring a small decrease in both mean error and RMSE for complex trajectories. Especially apparent was the use of the RS term together with the least force regularizer in dataset 4.
The impact of the RS term depends on the drone positions in image space. E.g., if the trajectory only spans a narrow horizontal strip, the inter-row delay will be small and its influence might be negligible. 
The interplay of the motion priors and the RS term is of interest for future investigation. 

On dataset 5, the results of the motion priors and RS correction were not entirely consistent with datasets 1-4, but due to the questionable accuracy of the ground truth for dataset 5 we do not consider these deviations significant.

Although not apparent in the statistical indicators, the motion regularizers and RS model locally improve parts of the trajectory significantly compared to the base solution with spline trajectory. In figure~\ref{fig:dataset5_diff} we highlight exemplary cases with blue circles where the trajectory with the full model is clearly more plausible.

\subsection{Robustness to de-synchronization}
One of the major advantages of our pipeline for practical purposes is that the input videos need not be synchronized.
Table ~\ref{tab:results_sync} summarizes our ablation study to test the effect of individual steps when the input sequences are out of sync.
Time shifts between input videos are modelled in two places: first, when fitting the two-view geometries (Sec.~\ref{sec:two-view}) we estimate a time shift $\beta_F$, then during bundle adjustment we refine it to obtain a final estimate $\beta_{opt}$.
The mean errors of these parameters are shown in the table for different amounts $\beta_{init}$ of desynchronization.
We add two baselines. The naive one completely ignores the time shift and assumes the cameras are in sync (or the impact of desynchronization is negligible).
The second one drops the time shift parameter only from the initial two-view fitting, thus testing the assumption that the initialisation will nevertheless be good enough to recover the time shift during bundle adjustment.

Table~\ref{tab:results_sync} shows that the the two-view time shift estimation is extremely robust and recovers sub-frame $\beta_F$ even with a large initial shift of 50 frames. The $\beta_{opt}$ after BA are only marginally better. The mean error of the reconstructed trajectory is not increased at all, i.e., our method fully compensates for the influence of desynchronization (the slight decrease with stronger de-sync is an artefact of the implementation, which then reconstructs slightly shorter trajectories). 
Without $\beta_F$, just optimizing $\beta_{opt}$ in the BA part of the pipeline, the mean reconstruction error and the resulting time shift error after BA immediately increase already at an initial time shift of 5 frames. For 10 frames shift, the pipeline fails, as the optimization gets stuck in a local minimum. For 15 frames shift the pipeline recovers the trajectory but with a large mean error of 83.8 cm and 7.7 frames mean error in the time shift. With neither $\beta_F$ nor $\beta_{opt}$, ignoring time shift altogether, the resulting trajectory has $>$1 m error already at 5 frames time shift, and fails for any larger shift.

\section{Conclusions}

We have described a practical method for reconstructing the trajectories of flying objects using multiple external cameras. Contrary to previous work, our method needs neither synchronized cameras nor pre-calibrated camera poses. Furthermore, it accounts for the influence of rolling shutters, and can handle gross errors of the preceding object detector.
We have shown that, thanks to these properties, a cheap and easy-to-use system can be built with off-the-shelf cameras, smartphones or surveillance cameras. Moreover, we have created a dataset with cm-accurate ground truth UAV trajectories, measured with differential GNSS. In our experiments we were able to achieve mean trajectory errors $<$40 cm relative to that ground truth. The dataset and the software pipeline are publicly available at \url{https://github.com/CenekAlbl/drone-tracking-datasets}, respectively \url{https://github.com/CenekAlbl/mvus}.
So far we have only explored the ability to reconstruct UAV trajectories in post-processing, after collecting all the data. A logical next step is to reconstruct the trajectory online, as video data comes in. Moreover, it would be interesting to implement multi-target tracking.



{\small
\bibliographystyle{IEEEtran}
\bibliography{references}
}

\end{document}